\documentclass[letterpaper, 10 pt, conference]{ieeeconf}  

\IEEEoverridecommandlockouts                              

\usepackage{graphics} 
\usepackage{times} 
\usepackage{amsmath} 
\usepackage{amssymb}  
\usepackage{graphicx}
\usepackage{color}
\usepackage{cite}
\makeatletter
\let\NAT@parse\undefined
\makeatother
\usepackage[colorlinks,linkcolor=blue, citecolor=blue, anchorcolor=blue]{hyperref}
\usepackage{multirow}

\title{\LARGE \bf
A Magnetic-Actuated Vision-Based Whisker Array for Contact Perception and Grasping}

\author{Zhixian Hu, Juan Wachs, and Yu She
\thanks{This material is partially based upon work supported by the National Science Foundation under Grant NSF NRI \#1925194, NSF \#2140612, NSF \#2423068, and Showalter Trust. Any opinions, findings, and conclusions or
recommendations expressed in this material are those of the author(s) and do not necessarily reflect the views of the funding agencies.}
\thanks{School of Industrial Engineering, Purdue University, West Lafayette, IN, USA. \tt\small{\{hu934, jpwachs, shey\}@purdue.edu}}
}

\begin{document}

\maketitle
\thispagestyle{empty}
\pagestyle{empty}

\begin{abstract}
Tactile sensing and the manipulation of delicate objects are critical challenges in robotics. This study presents a vision-based magnetic-actuated whisker array sensor that integrates these functions. The sensor features eight whiskers arranged circularly, supported by an elastomer membrane and actuated by electromagnets and permanent magnets. A camera tracks whisker movements, enabling high-resolution tactile feedback. The sensor's performance was evaluated through object classification and grasping experiments. In the classification experiment, the sensor approached objects from four directions and accurately identified five distinct objects with a classification accuracy of 99.17\% using a Multi-Layer Perceptron model. In the grasping experiment, the sensor tested configurations of eight, four, and two whiskers, achieving the highest success rate of 87\% with eight whiskers. These results highlight the sensor's potential for precise tactile sensing and reliable manipulation.

\end{abstract}

\section{INTRODUCTION}

Tactile sensing is essential in robotic systems, providing robots with the ability to physically perceive and interact with the environment. Objects can be characterized through contacts, and shape and texture are found through tactile interaction. One promising direction in the broad area of tactile sensing is the use of artificial whiskers. Such end-effectors draw inspiration from the vibrissae (whiskers) of animals such as rodents and seals \cite{yu2024whisker}. In nature, biological whiskers are highly sensitive mechanoreceptors which enable animals to detect subtle environmental changes through physical contact and airflow \cite{boubenec2012whisker}. When used in combination with motion, creatures can explore and making useful decisions about their surroundings. In robotic systems, whisker-based sensors mimic these biological systems, providing adaptable tactile feedback that is well-suited for tasks that require delicate object detection and manipulation.

\begin{figure}[thpb]
  \centering
  \includegraphics[scale=0.8]{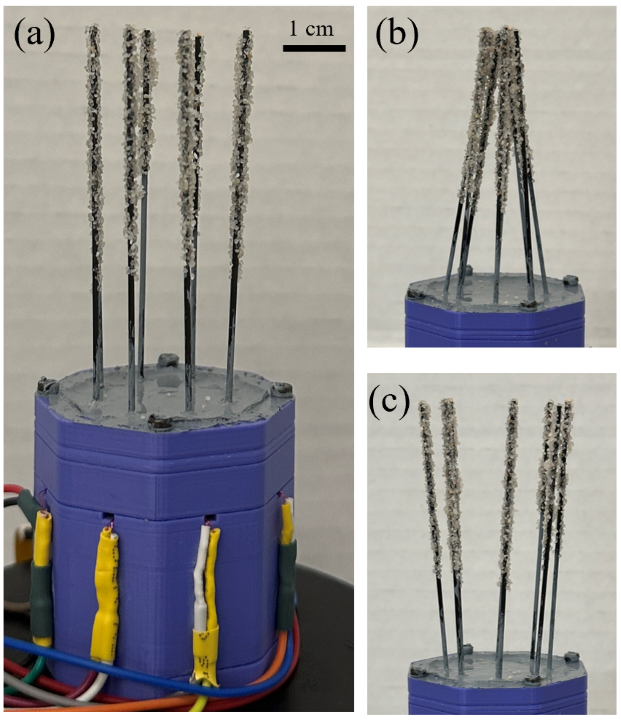}
  \caption{Overview of the vision-based magnetic-actuated sensor. (a) Neutral state. (b) Retraction state. (c) Extension state.}
  \label{intro}
\end{figure}

Current whisker-based tactile sensors utilize modalities such as magnetic\cite{lin2022whisker,kossas2024whisker,wegiriya2019stiffness}, piezoresistive\cite{zhang2022small}, piezoelectric\cite{wegiriya2019stiffness, wang2023tactile}, triboelectric\cite{liu2022whisker}, and barometer-based\cite{ye2024biomorphic, xiao2022active}. While these approaches are effective in measuring forces, displacements, and contact interactions, they often involve trade-offs regarding sensitivity, operational range, and overall system complexity. One challenge with these sensors is the difficulty in discerning the direction of contacts, which is crucial for accurately interpreting interactions. To overcome this limitation, various strategies have been proposed, such as embedding multiple sensing elements\cite{zhang2022small, gul2018fully, emnett2018novel}, using arrays of whiskers\cite{xiao2022active, deer2019lightweight}, or magnetic-based terminals \cite{yu2022bioinspired, lin2022whisker}. While these strategies enhance directional sensitivity, they also increase system complexity, due to associated technical challenges in integration and calibration processes.

Vision-based whisker systems address key limitations of traditional tactile sensing by effectively detecting the direction of contact. For example, optical sensors arranged in particular geometric arrays can allow simultaneously sensing through multiple whisker movements  \cite{kent2021whisksight,lepora2018tacwhiskers}. This is in contrast to magnetic-based methods, which struggle with array integration due to interference and complexity in handling individual magnetic fields\cite{zhou2022magnetized}. Vision-based approaches use cameras or optical sensors to capture the movements of all whiskers in real time. These sensor arrays allow for the simultaneous detection of multiple contact points, enabling robots to map their surroundings more comprehensively. Vision-based sensing allows direct monitoring of whisker movements without adding additional mechanical complexity to such systems.

While traditional static whisker sensors are effective in detecting the presence and texture of objects, they are inherently limited in their ability to interact with and manipulate these objects. To address this issue, it has been proposed attaching rotational motors at the base of the whiskers, enabling sweeping motions for active sensing \cite{sullivan2011tactile, assaf2016visual, xiao2023complacent}. Additionally, electromagnetic actuation using a permanent magnet at the whisker root actuated by nine surrounding electromagnetic coils was proposed \cite{yu2022bioinspired}. In \cite{lepora2018tacwhiskers}, a whisker array with a central tendon mechanism that protracts and retracts the whiskers to achieve active whisking motion was shown. However, these methods often result in systems that are bulky and complex due to the large size of the actuators, or lack of controllability, eventually affecting their effectiveness and flexibility.

In this paper, we present a novel vision-based whisker sensor array designed for delicate object manipulation and environmental sensing, as shown in Fig. \ref{intro}. Our sensor array features eight whiskers arranged in a circular configuration, with each whisker capable of independent forward and backward magnetic actuation. This design enables the sensor array to grasp and manipulate small, light, delicate objects, a capability that is not feasible with traditional static whisker systems. The vision-based component of our array allows for accurate monitoring of whisker deflections, enabling the system to distinguish between contact directions and forces applied from multiple angles. This feature is particularly valuable in applications that require precise control, such as soft robotics, biomedical devices, and micro-manipulation.

The remainder of this paper is structured as follows: Section 2 discusses the design and construction of the whisker sensor array, including the vision-based sensing mechanism and actuation system. Section 3 presents experimental results demonstrating the sensor's ability to detect and classify objects, and grasp delicate objects. Finally, Section 4 concludes with a discussion of future directions.

\section{METHOD}
The system is a vision-based whisker array with magnetic actuation, integrating both tactile sensing and object manipulation capabilities. The system comprises eight whiskers supported by an elastomer membrane. Each whisker has an electromagnet at its base, paired with corresponding permanent magnet for actuation. A camera is below the base and used for tracking all whisker movements, complemented by a light source. This design provides simultaneous multi-whisker sensing and actuation.

\subsection{Mechanical Design}
\begin{figure}[thpb]
  \centering
  \includegraphics[scale=0.45]{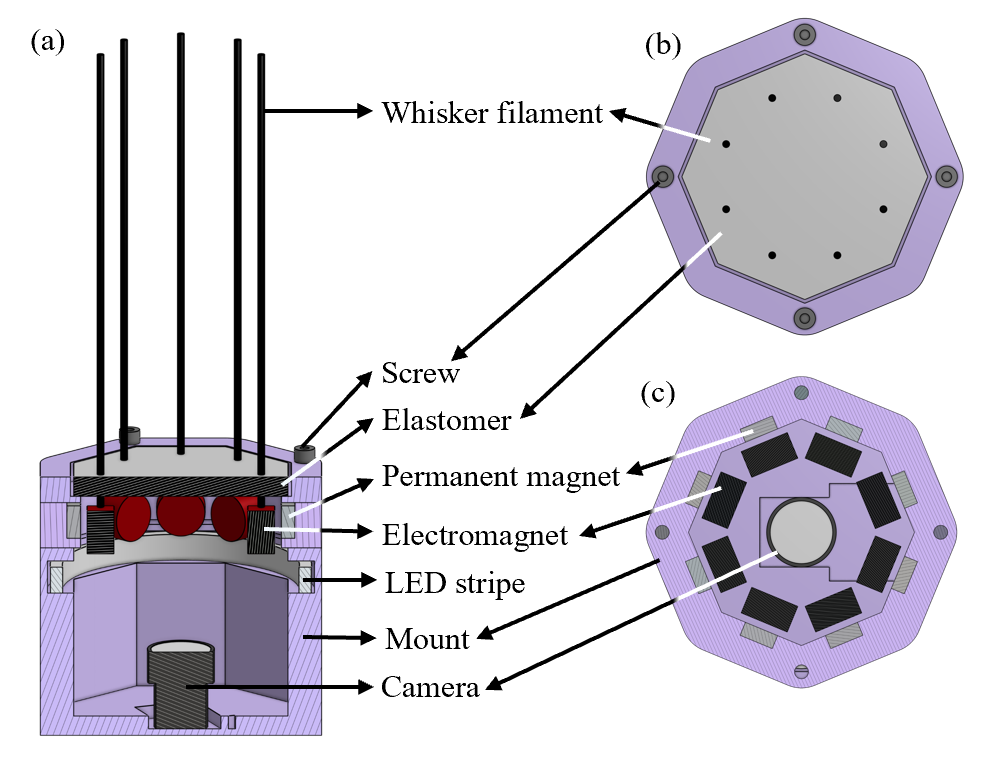}
  \caption{Mechanical design of the sensor. (a) Overview of the sensor's mechanical design. (b) Top view of the sensor. (c) Top cross-section view showing the arrangement of the electromagnets and permanent magnets.}
  \label{mechanicalDesign}
\end{figure}

The mechanical design centers around a vision-based whisker array, as shown in Fig. \ref{mechanicalDesign} (a). The array consists of eight whiskers uniformly arranged in a circular layout with a 12 mm radius (shown in Fig. \ref{mechanicalDesign} (b)), all supported by a single elastomer membrane made of Ecoflex 00-30 (Smooth-On, Inc.) with a thickness of 3 mm. The membrane has an octagon shape, with an inscribed circle of 16 mm radius. The selections of the membrane material and thickness were determined empirically through experimentation to provide the sufficient flexibility and structural support.  To minimize the effects of external lighting on tactile perception, the upper surface of the membrane is painted with gray ink.

The whiskers pass through the membrane and are glued to secure them in place. Each whisker filament is a carbon fiber rod with a 1 mm diameter and a total length of 69 mm long, of which 63 mm extends above the membrane. Inspired by the structure of insect legs, the tops of the whiskers are coated with sand to create a spongy layer that increases friction. Controlled trials verified the effectiveness of this modification, demonstrating that the sensor can grasp light objects with the sand that it cannot grasp without it. Each whisker is equipped with a disc electromagnets, measuring 4 mm in thickness and 4 mm in diameter, which faces a fixed permanent magnet mounted on the base (illustrated in Fig. \ref{mechanicalDesign} (c)). The permanent magnet provides a magnetic field for the electromagnet, and when the electromagnet is powered, magnetic forces push it towards or away from the permanent magnet, actuating the whisker.

A LED stripe is positioned beneath the permanent magnets on the mount to serve as a light source for the camera. A Raspberry Pi camera with a 160° variable focus lens is used to capture whisker movements, connected to a Raspberry Pi 4 Model B for image streaming and processing. The camera's focus is adjusted to the position of the electromagnets, which are tracked for tactile perception. A custom mount, 3D printed with PLA material, is designed to hold all components together securely.

\begin{figure*}[thpb]
  \centering
  \includegraphics[scale=0.8]{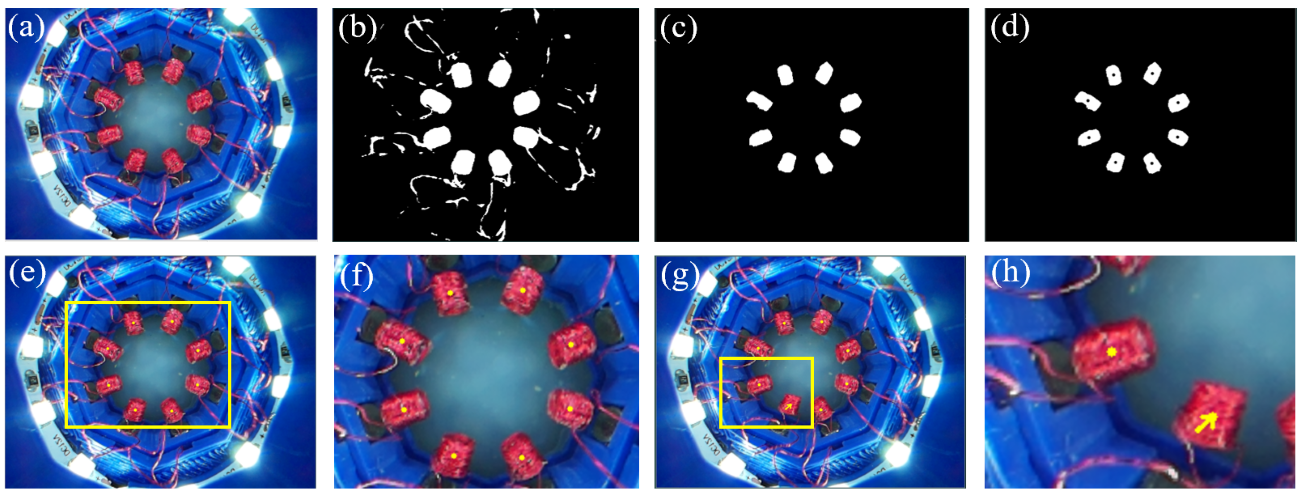}
  \caption{Image processing pipeline of the whisker tracking system. (a) Raw image. (b) Mask showing the filtered red color region. (c) Eroded image to eliminate the effects of the wires. (d) Detection of contour centers. (e) Image with yellow dots indicating the detected tracker centers when no external forces are applied to the whiskers. (f) Zoomed-in view of the yellow rectangle from (e). (g) Image with yellow arrows showing the movement of the tracker centers when external forces are applied to the whiskers. (h) Zoomed-in view of the yellow rectangle from (g)}
  \label{imgPrc}
\end{figure*}

\subsection{Whisker Tracking}

The whisker tracking system is designed to capture simultaneous movement of all eight whiskers, enabling precise tactile sensing. A camera is positioned beneath the whisker array to monitor the real-time positions of the electromagnets. The electromagnets act as visual markers, allowing the camera to track whisker movements with high spatial resolution. A dedicated light source illuminates the array to enhance contrast and ensure reliable detection of the markers under various lighting conditions.

Image processing algorithms are utilized to analyze the camera feed, extracting position and movement data for each whisker. The raw images, captured at a resolution of 640$\times$480 pixels (Fig. \ref{imgPrc} (a)), are first converted into the HSV (hue, saturation, and value) color space to facilitate more robust color-based segmentation. A mask is then applied to isolate the red-colored areas corresponding to the electromagnets (Fig. \ref{imgPrc} (b)). To mitigate the influence of red wires associated with the electromagnets, erosion is performed using a 10×10 kernel, effectively reducing noise and enhancing marker detection accuracy (Fig. \ref{imgPrc} (c)). Contour detection is subsequently carried out using the 
\texttt{findContours} function \cite{suzuki1985topological} in Python’s OpenCV library \cite{bradski2000opencv} to locate the electromagnet markers. The centroid of each detected contour is then
determined from the first-order and zero-order image moments by using the \texttt{moments} function (Fig. \ref{imgPrc} (d)), allowing for precise tracking of the electromagnet centers (Fig. \ref{imgPrc} (e)).

This tracking pipeline enables the system to compare the positions of the electromagnet centers under different conditions: when no external forces are applied to the whiskers and when external forces, including magnetic forces for actuation, are present. By analyzing these positional differences, the direction and magnitude of whisker movements can be inferred. It is assumed that the pivot of the whisker movement is located at the center of the membrane. The length of the whisker below the pivot is denoted as $l_l$, and the length above the membrane is denoted as $l_u$. Given the known distance between the base of the whisker and the camera, the spatial resolution of the camera (distance per pixel) can be determined, representing by $k$. Since the movement of the whisker base is small, variations in the distance per pixel across the camera image can be considered negligible. Thus, given the pixel displacements  $p_x$ and $p_y$ of the electromagnet centers, the deflection angle $\theta$ and the displacement of the whisker tip can be computed as follows:
\begin{equation}\label{angle}
\theta = arcsin(\frac{k\sqrt{p_x^2+p_y^2}}{l_{l}})
\end{equation}
\begin{equation}\label{x}
x_{tip} = -\frac{l_{u}}{l_{l}}kp_x,
\end{equation}
\begin{equation}\label{y}
y_{tip} = -\frac{l_{u}}{l_{l}}kp_y,
\end{equation}
\begin{equation}\label{z}
z_{tip} = l_{u} cos\theta,
\end{equation}
where $\theta$ is defined as the angle between the current whisker position and its neutral state, and the coordinates $x_{tip}, y_{tip}$ and $z_{tip}$ refer to the tip position of the whisker relative to the original point of the whisker movement pivot. 

This vision-based approach provides a comprehensive method for detecting whisker movement, overcoming the limitations of traditional tactile sensors that rely on discrete sensing elements for each whisker. It is assumed that the carbon fiber whiskers behave as rigid bodies during interactions with the external environment, simplifying the analysis of movement dynamics.

\subsection{Whisker Actuation}

\begin{figure*}[thpb]
  \centering
  \includegraphics[scale=0.35]{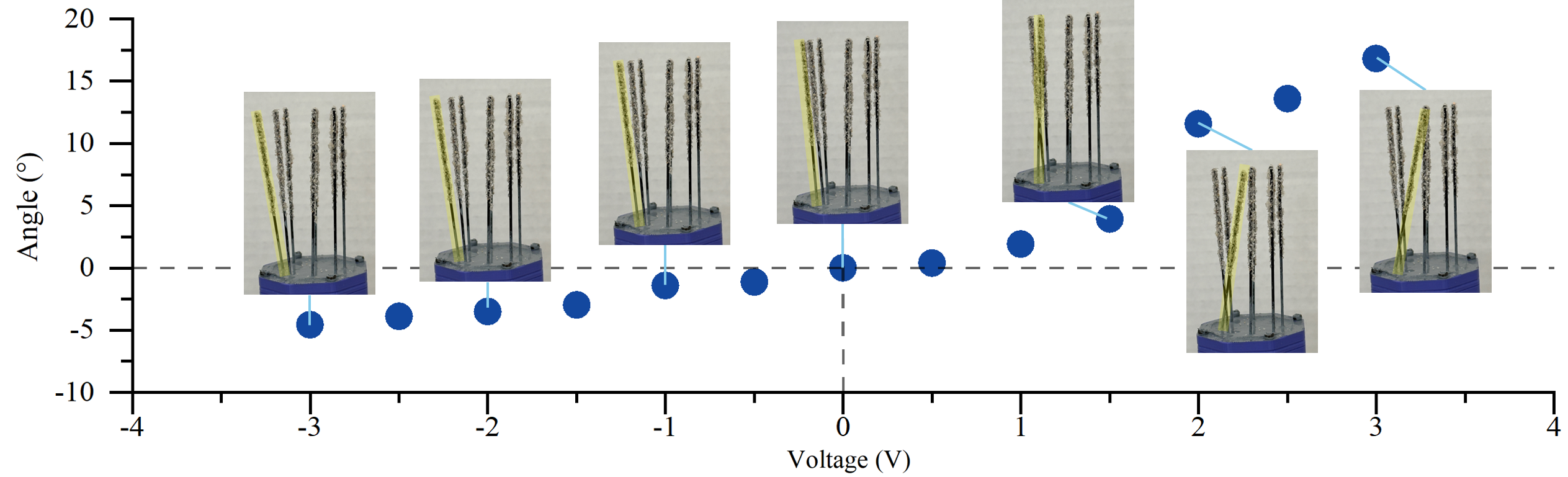}
  \caption{Deflection angles of a whisker at various electromagnet voltages. A zero angle represents the neutral state, positive values indicate retraction, and negative values indicate extension.}
  \label{votAngle}
\end{figure*}

Whisker actuation is achieved through the interaction between the electromagnets located at the bases of the whiskers and corresponding permanent magnets. Each electromagnet is constructed using enameled copper wire with a diameter of 0.101 mm, wound into approximately 600 turns, resulting in an electromagnetic coil about 4 mm in diameter and 4 mm in height. The resistance of each electromagnet is approximately 7 $\Omega$. Neodymium permanent magnets, each with a height of 1.59 mm and a diameter of 4.76 mm, orient with the ``N" pole facing the electromagnet to provide a consistent magnetic field. By controlling the magnitude and polarity of the electromagnetic field, the whiskers can be actuated to move either towards the center of the array or away from it, facilitating retraction and extension actions of the whisker array. The deflection angles of a whisker at various electromagnet voltages are shown in Fig. \ref{votAngle}, where a zero angle represents the neutral state, positive values indicate retraction, and negative values indicate extension. 

Each electromagnet is connected to a dual H-bridge motor drive module (L298N), which enables bidirectional control of current flow. The output voltage from the L298N is further connected to a voltage dividing circuit, which regulates the voltage supplied to the electromagnets, allowing for different operational voltage levels to be applied as needed. The L298N modules are interfaced with the Raspberry Pi, which handles the control signals for the actuation of each whisker unit independently, allowing for individual adjustment of polarity. This approach aims to bridge the gap between tactile sensing and active interaction, potentially enabling more complex and adaptive behaviors in robotic applications.

\section{EXPERIMENTS}
To verify the performance of the vision-based whisker array sensor, two experiments were conducted: an object classification experiment and an object grasping experiment. The first experiment aimed to evaluate the sensor's ability to classify objects based on tactile feedback, while the second experiment tested the sensor's capability to grasp light objects using different numbers of whiskers. In both experiments, the sensor was mounted on a UR 16e industrial robot arm from Universal Robots, which provided precise movement and positioning capabilities. This setup allowed for controlled and repeatable interactions between the sensor and the test objects.

\subsection{Object Classification Experiment}

\begin{figure}[thpb]
  \centering
  \includegraphics[scale=0.8]{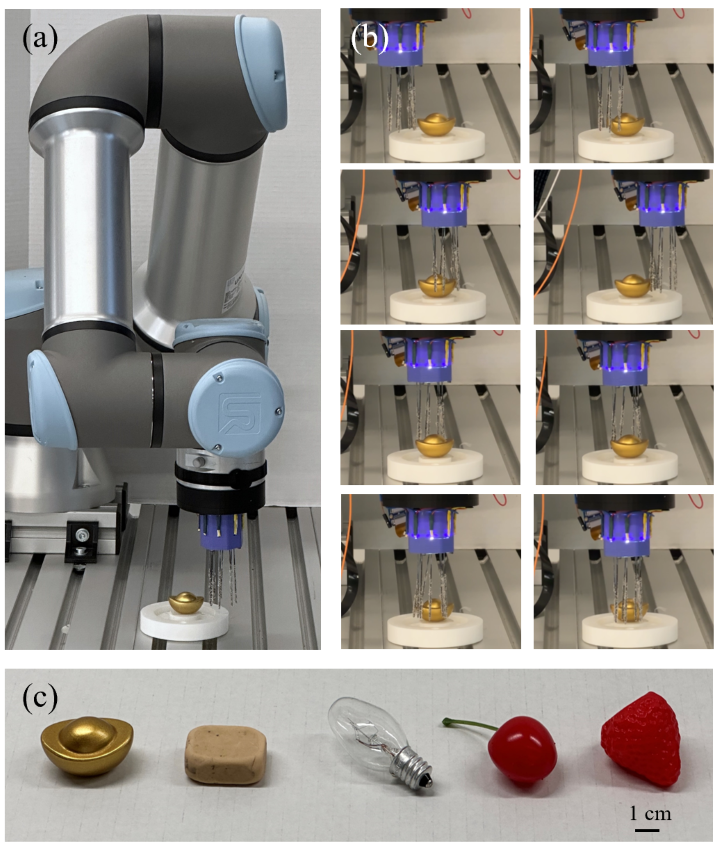}
  \caption{(a) Experiment setup of the object classification experiment. (b) The sensor contacts the object from four directions at varying distances. (c) The five tested objects: yuanbao, eraser, light bulb. plastic cherry, and plastic strawberry.}
  \label{objCls}
\end{figure}
\begin{figure}[thpb]
  \centering
  \includegraphics[scale=0.6]{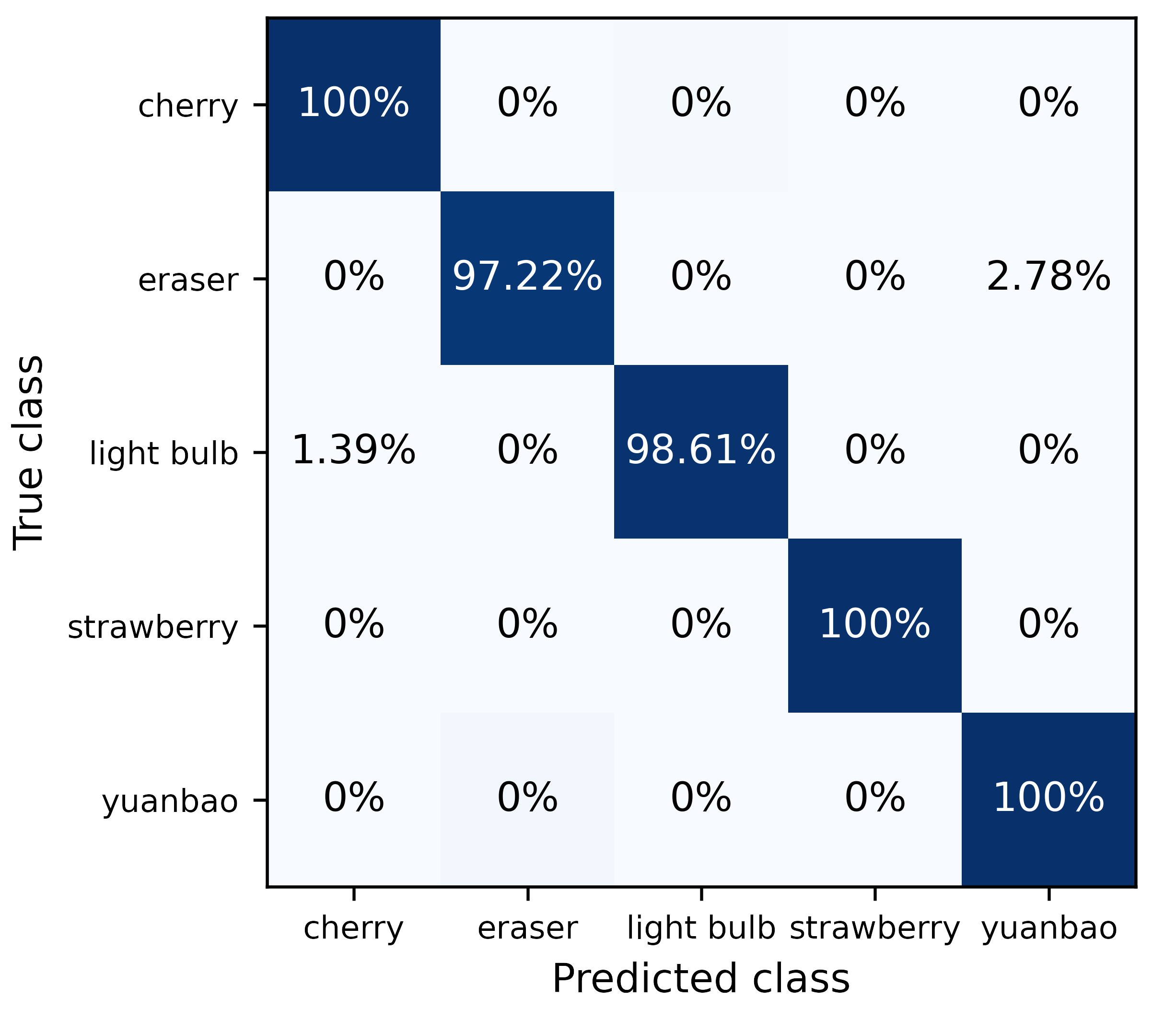}
  \caption{Confusion matrix of the object classification experiment.}
  \label{confusion}
\end{figure}

The object classification experiment (Fig. \ref{objCls} (a)) was designed to assess the sensor’s ability to distinguish between different objects based on tactile interactions. As shown in Fig. \ref{objCls} (c), five distinct objects were selected for this experiment, varying in shape to provide various tactile feedback. The sensor, mounted on the UR16e robot arm, approached each object horizontally from four different directions. Data was collected at distances of 3 cm and 1 cm from the object center, as illustrated in Fig. \ref{objCls} (b). The vertical gap between the whisker tips and the surface on which the objects were placed was approximately 3 mm. For each trial, the sensor collected tactile data points by recording the movements of all eight whiskers using the whisker tracking method, resulting in eight contact data points per approach. To ensure comprehensive data coverage, the sensor was rotated from 0° to 300° in steps of 60° and then approached the object, with five trials conducted at each rotation angle, resulting in 30 samples per object. To further augment the dataset, the collected whisker movement images were rotated by 45° to simulate additional contact scenarios, as each whisker is considered equivalent in the classification task. This augmentation increased the dataset to 240 samples per object, totaling 1200 samples across all objects. The dataset was then split into training and testing subsets in a 7:3 ratio.

A Multi-Layer Perceptron (MLP) model was employed for the classification task, featuring one hidden layer with 32 neurons and ReLU activation functions. The model was trained using the Adam optimizer with a learning rate of 0.01, and cross-entropy loss was used as the criterion. The training was conducted over 10 epochs. A classification accuracy of 99.17\% was achieved on the test set, demonstrating the sensor’s high capability for object identification based on tactile profiles. The confusion matrix, shown in Fig. \ref{confusion}, illustrates the classification performance across the different objects. The matrix reveals that a small number of eraser samples were misclassified as yuanbao, likely due to similar size and shape. Several light bulb samples were misclassified as cherries, potentially due to the similar size and shape, along with the presence of the cherry stem. Overall, the high classification accuracy indicates the effectiveness of the sensor in distinguishing between objects, validating its potential for advanced tactile sensing applications.

\subsection{Object Grasping Experiment}
\begin{figure}[thpb]
  \centering
  \includegraphics[scale=0.65]{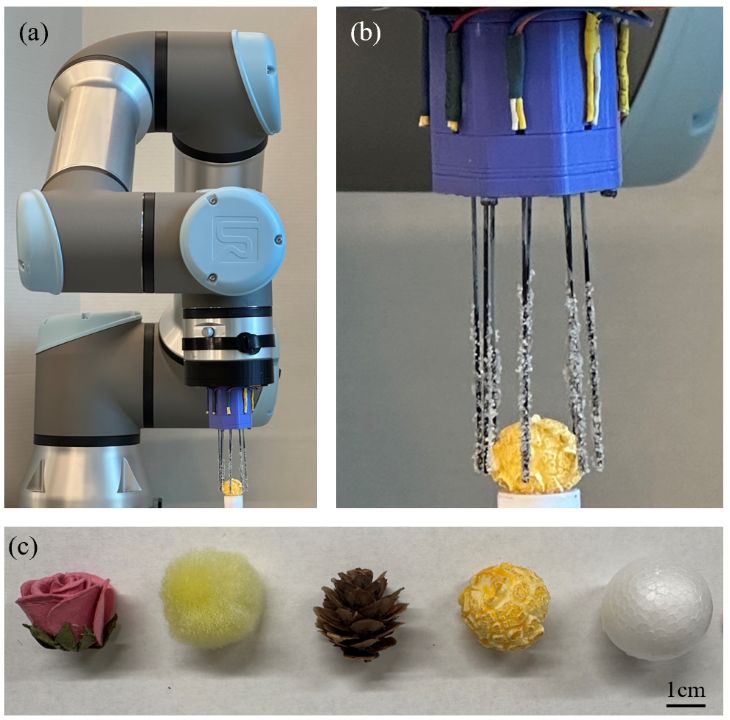}
  \caption{(a) Experiment setup of the object grasping experiment. (b) Zoomed-in view of (a) showing the sensor and the tested object. (c) The five tested objects: paper flower, pom-pom, mini pinecone, popcorn, and foam ball (from left to right).}
  \label{expGrasp}
\end{figure}

\begin{figure}[thpb]
  \centering
  \includegraphics[scale=0.6]{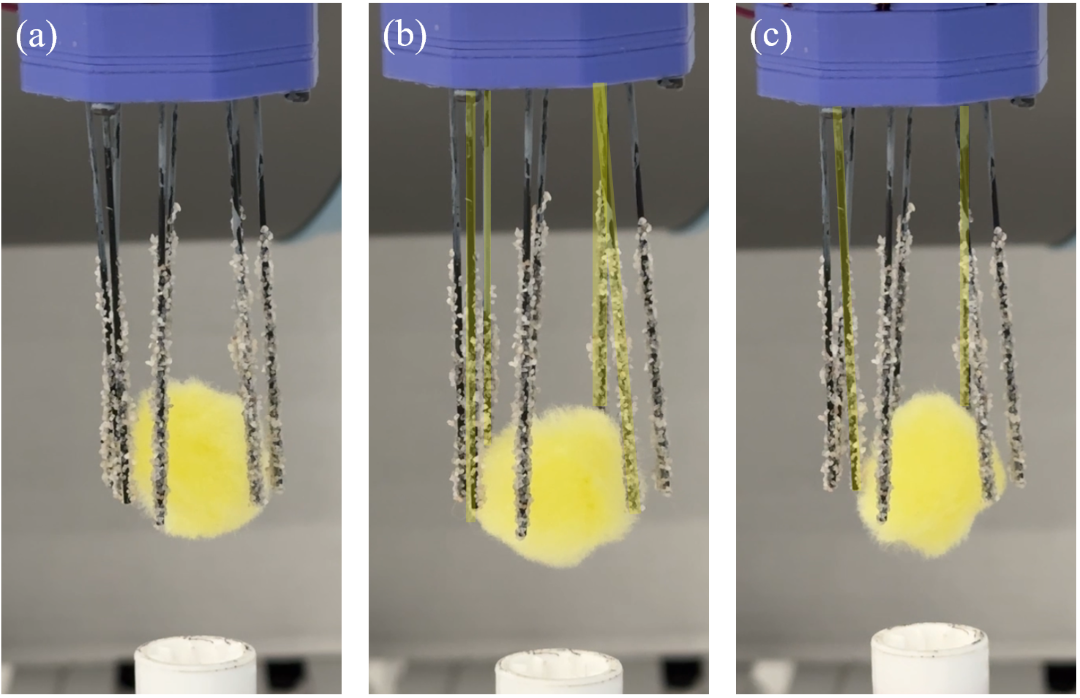}
  \caption{The whisker sensor array grasping the pom-pom using (a) eight whiskers, (b) four whiskers, and (c) two whiskers.}
  \label{expGraspWsk}
\end{figure}

\begin{table}[h]
\caption{Grasp Success Rate of The Three Whisker Configuration for the Five Objects (\%)}
\label{gspSuccessRate}
\begin{center}
\begin{tabular}{|c|c|c|c|}
\hline
~&Eight Whiskers & Four Whiskers & Two Whiskers\\
\hline
Paper flower & 70 & 35 & 20\\
\hline
Pom-Pom & 100 & 100 & 65\\
\hline
Mini Pinecone & 80 & 50 & 30\\
\hline
Popcorn & 85 & 35 & 10 \\
\hline
Foam Ball & 100 & 50 & 30 \\
\hline
Total & 87 & 54& 31\\
\hline
\end{tabular}
\end{center}
\end{table}

The object grasping experiment aimed to evaluate the sensor’s effectiveness in grasping light objects using varying numbers of whiskers. As illustrated in Fig. \ref{expGrasp} (a), the sensor was mounted on the UR16e robot arm, which controlled the sensor’s movement for precise and repeatable grasp attempts. Five light objects with different shapes and surface properties were chosen for this experiment, which were paper flower, pom-pom, mini pinecone, popcorn, and foam ball (shown in Fig. \ref{expGrasp} (c)). The weights of these objects were 0.97 g, 0.33 g, 0.82 g, 0.36 g, and 0.15 g, respectively.

For each grasp attempt, the sensor was positioned above the object, with the object roughly centered beneath the sensor. The UR16e robot arm then moved the sensor vertically downwards until it reached the lowest point, where the vertical space between the whisker tips and the surface on which the object was placed was approximately 2 mm. At this position, the sensor’s whiskers were actuated to grasp the object. The actuation voltage for each electromagnet was 3V. Following actuation, the robot arm lifted the sensor upwards by 3 cm and held this position briefly. A grasp was considered successful if the object remained held by the sensor without falling.

\begin{figure}[thpb]
  \centering
  \includegraphics[scale=0.5]{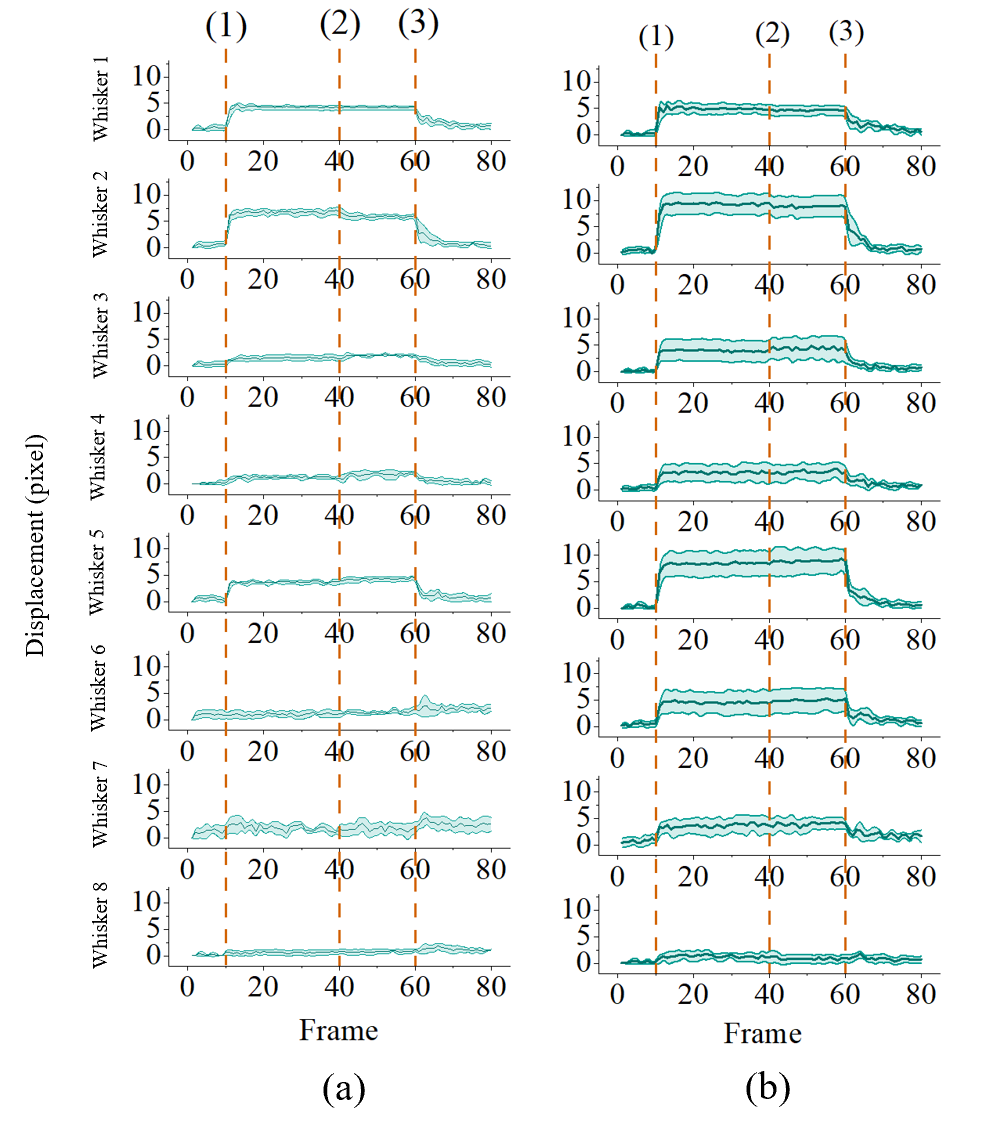}
  \caption{Mean displacement and error bands of each whisker over time across five trials during the object grasping experiment for (a) the foam ball and (b) the mini pinecone. Key phases are marked: (1) transition from neutral to retraction state, (2) robot arm moving upward, and (3) whiskers returning from retraction to neutral state. }
  \label{repeat}
\end{figure}

For each object, the sensor performed 20 grasp attempts using a single whisker configuration at a time. Three whisker grasping configurations were tested: grasping with all eight whiskers, four whiskers and two whiskers. Example grasps for each configuration are shown in Fig. \ref{expGraspWsk}. The success rate of each configuration for each object was recorded, as illustrated in Table \ref{gspSuccessRate}. The configuration using all eight whiskers achieved an average success rate of 87\%, with the minimum success rate being 70\%. This configuration consistently exhibited the highest success rates across all objects, highlighting its superior stability and control in grasping tasks.

The variation in grasp success rates among the tested objects can be attributed to differences in object configuration, weight distribution, and surface properties. For the eight-whisker configuration, failures were primarily due to slippage between the whisker rods and the objects. Objects with higher surface friction, such as the pom-pom with its furry exterior, achieved a 100\% success rate due to increased friction between the whisker rods and the object. Similarly, the foam ball also achieved a 100\% success rate, attributed to its uniform weight distribution and light weight, which promoted stable grasping. In contrast, the four-whisker configuration encountered reduced friction between the whiskers and the objects, posing additional challenges for stable grasping. Despite these challenges, this configuration still achieved an average success rate of over 50\%, demonstrating its potential for grasping under less favorable conditions. The two-whisker configuration required precise alignment, with the forces exerted by the whiskers needing to pass through the object's center of mass and the friction being sufficient to maintain the grasp. Due to the reduced fault tolerance, this configuration resulted in an average success rate of 31\%, which, while lower, remains acceptable given the constraints. 

It is noteworthy that all grasping trials were conducted without force control, and all objects were preserved intact during and after the grasping attempts. This outcome underscores the whisker array's capability to manipulate delicate objects without causing damage, demonstrating its potential advantages in applications that require gentle handling.

Furthermore, recorded frames of the foam ball and the mini pinecone were analyzed to evaluate the performance of the whisker actuation and tactile sensing. The displacement of each whisker was plotted in Fig. \ref{repeat} against the frame number, producing a mean and error band plot for each object, with data aggregated from five trials using eight whiskers per object. The plot illustrates three key phases: (1) the transition command from the neutral state to the retraction state for the whiskers, (2) the command for the robot arm to move upward, and (3) the command for the whiskers to return from the retraction state to the neutral state. The displacement data for the foam ball demonstrate the reliability and consistency of the whisker actuation, with minimal variations across trials, indicating stable performance. In contrast, the data for the mini pinecone show greater variability, reflecting the whiskers' sensitivity to the differences in object characteristics such as shape and orientation. This comparison highlights the sensor's ability to perceive and respond to variations in manipulated objects, as evidenced by the larger displacement variations for the pinecone compared to the foam ball. These findings suggest that the whisker array is capable of distinguishing between different objects based on tactile feedback, which is critical for advanced manipulation tasks in robotics.

\section{CONCLUSION}

This study presented a novel vision-based magnetic-actuated whisker array sensor that integrates tactile sensing and object manipulation to address limitations in existing tactile sensors. The sensor’s design, featuring a circular array of eight whiskers actuated by electromagnets, enables precise control over whisker movements and high-resolution tactile feedback through a camera-based tracking system. The object classification experiment demonstrated the sensor’s ability to accurately distinguish between five objects with varying shapes and materials, which highlights the sensor's potential for applications in object characterization. Additionally, the object grasping experiment showed that the eight-whisker configuration provided superior stability and control in grasping tasks, confirming the sensor’s effectiveness in advanced tactile sensing and manipulation, making it a promising tool for robotics and automated systems.

Future work will focus on integrating sensing and actuation to enable delicate manipulation with real-time feedback, enhancing the sensor’s ability to perceive object and environmental states during interactions. Efforts will also include refining the sensor's sensitivity to detect finer tactile interactions and improving the precision of whisker movement control through advanced voltage regulation. Addressing the overheating issue of the electromagnets will be a critical priority to ensure reliable and sustained performance. These advancements aim to further enhance the sensor’s versatility and effectiveness in complex and dynamic environments.

\addtolength{\textheight}{-10.5cm}   





\bibliographystyle{IEEEtran}
\bibliography{IEEEexample}

\end{document}